\documentclass[journal,transmag]{IEEEtran}

\usepackage{systeme}
\usepackage{xcolor}
\usepackage{colortbl}
\usepackage{lipsum}
\usepackage{placeins}
\usepackage{longtable}
\usepackage{soul}

\ifCLASSINFOpdf
\else
\fi
\usepackage[cmex10]{amsmath}
\usepackage{algorithmic}

\usepackage{array}

\usepackage{stfloats}
\begin{document}
%
\title{Wind ramp event prediction with parallelized Gradient Boosted Trees}


\author{\IEEEauthorblockN{Saurav Gupta\IEEEauthorrefmark{1},
Nitin Anand Shrivastava\IEEEauthorrefmark{1},
Abbas Khosravi\IEEEauthorrefmark{2},
Bijaya Ketan Panigrahi\IEEEauthorrefmark{1},~\IEEEmembership{Senior Member,~IEEE}}
\IEEEauthorblockA{\IEEEauthorrefmark{1}Department of Electrical Engineering, Indian Institute of Technology Delhi, India}
\IEEEauthorblockA{\IEEEauthorrefmark{2}Center for Intelligent Systems Research, Deakin University, Australia}
\thanks{Saurav Gupta is with the Department of Electrical Engineering, Indian
Institute of Technology, New Delhi, India. (e-mail: sauravguptaaa@gmail.com)
Nitin Anand is with the Department of Electrical Engineering, Indian
Institute of Technology, New Delhi, India. (e-mail: anandnitin26@gmail.com)
A. Khosravi is with the Center for Intelligent Systems Research, Deakin
University, Australia.(e-mail:abbas.khosravi@deakin.edu.au)
B.K.Panigrahi is with the Department of Electrical Engineering,
Indian Institute of Technology, New Delhi, India. (email:
bijayaketan.panigrahi@gmail.com.)}}

\markboth{Journal of \LaTeX\ Class Files,~Vol.~13, No.~9, September~2014}%
{Shell \MakeLowercase{\textit{et al.}}: Bare Demo of IEEEtran.cls for Journals}
\IEEEtitleabstractindextext{%
\begin{abstract}
Classifying the  wind power ramp events. Accurate prediction of wind ramp events is critical for the ensuring the reliability and stability of the power systems with high penetration of wind energy. This paper proposes a classification based approach for estimating the future class of wind ramp event based on certain thresholds. A parallelized gradient boosted regression tree based technique has been proposed to accurately classify the normal as well as rare extreme wind power ramp events. The model has been validated using wind power data from Alberta electricity market. Performance comparison with several benchmark techniques indicates that the model superiority of the proposed technique in terms of superior classification accuracy. 
\end{abstract}

}

\maketitle

\IEEEdisplaynontitleabstractindextext

%
\IEEEpeerreviewmaketitle

\section{Introduction}
%
%
%
%
\IEEEPARstart{G}{lobal} warming and depleting fossil fuel resources are some of the the key driving factors for the increasing popularity of renewable energy sources such as wind energy~\cite{Bilgen2008372}. Wind energy is a clean and sustainable alternate to the conventional sources of energy. Many countries have resolved to gradually increase the penetration of wind power into their existing power  systems~\cite{Doe:2009:Online}\cite{GWEO}\cite{MNRE}. However, the intermittent  nature of wind energy poses certain technical and economic challenges. Sudden increase or decrease of the wind energy in response to changes in wind speed, denoted as a wind ramp event, leads to challenges in maintaining the demand-supply balance at all times.   

Accurate forecast of wind power plays a key role in managing the uncertainty of wind power generation for its reliable integration with the grid. Wind power forecasts are employed in power system planning activities such as unit commitment, scheduling and dispatch by system operators. Several research papers have investigated the problem of accurate wind power forecasting and developed efficient and reliable methodologies. These models are based on physical, statistical and hybrid approaches. Some of the existing wide scale wind farm models are Prediktor, HIRPOM, ForeWind and OMEGA~\cite{WE:WE96},\cite{Lange}.  
 
 The existing wind energy forecasting literature primarily pertains to point forecast where the focus is one predicting the numerical value of the wind power at a given instant in future~\cite{Suganthan}\cite{Mohandes2004939}\cite{Hui2012545}\cite{Sfetsos200023}. Point forecasting techniques suffer with the limitation that they do not indicate the uncertainty associated with such forecasts. Since wind power is a highly uncertain variable, it is necessary to quantify the uncertainty associated with the forecasts along with their numerical values. Some papers address this issue by estimating the probability density function, confidence interval and prediction interval of point forecasts\cite{Pinson2007,Khosravi2013,Bludszuweit2008,Argonne,Bremnes2006}
 

Keeping in view the increasing capacity as well as penetration of the wind power into the existing power systems, it is imperative to accurately predict the severe ramp up/ ramp down events over a short period of time to ensure the stability and the reliability of the system. A survey of the available literature indicates the attempts made by reserachers to address the problem of predicting wind ramp events. A dynamic programming recursion technique was proposed by Sevlian and Rajagopal~\cite{sevlian6555972} to detect the ramp events. A family of scoring functions with ramp events definitions was suggested which could be used in forecasting and simulation. Bossavy et al.\cite{bossavy2010forecasting} generated the confidence intervals associated with the occurrence of ramp events. However, it is difficult to incorporate these intervals in automated dispatching and unit commitment systems. Graves et al.\cite{greaves2009temporal} suggested the incorporation of the numerical weather forecast (NWP) models for improving the forecasting accuracy of ramp events. But the success of their proposed technique depends on improving the NWP forecasts which is a difficult task. Cui et al.\cite{Mingjian7018976} proposed a method which provided the statistical information associated with wind power ramping events.      

\par Some alternative approaches have been proposed where the focus is on predicting the class of the ramp event based on certain  predefined-thresholds. Zareipour et al.\cite{Zarei6039625} predict the class of the ramp event using the one against all approach of the support vector machines (SVM). The paper reported satisfactory accuracy levels for different testing periods. But the paper lacks a formal structure for future control application~\cite{Mingjian7018976} and also a detailed analysis and specifications of the classification methodology employed. The wind power data is labeled for classification purpose based on predetermined thresholds. Since extreme ramp events are rare in nature, this leads to a huge class imbalance in the datasets leading to doubtful conclusions about the accuracy of the methodology.

\par This paper proposes a novel methodology based on gradient based regression trees to efficiently address such issues. The rest of the paper is organized as follows. The problem definition is discussed in Section II. The detailed methodology is presented in Section III. Section IV describes the various case studies performed to validate the methodology. Results and discussions are provided in Section V followed by the conclusions in Section VI.             

\section{Problem Definition}
Let the wind power at time t and $t-\Delta t$
of the d-day be denoted by $w_{d}(t)$ and $w_{d}(t-\Delta t)$, respectively. In this formulation, $\Delta t$ is a relatively small time interval (ranging from 10 minutes to 1 hour). 
The difference between the two consecutive wind power outputs is defined as $\Delta w_{d}(t)$ = $w_{d}(t)- w_{d}(t- \Delta t)$. 

A positive value for $\Delta w_{d}(t)$ is an increase in the wind power output i.e. a "ramp up" event while a negative value of $\Delta w_{d}(t)$ represents a decrease in the wind power output, which is a "ramp down" event. A ramp event is defined as $\Delta w_{d}(t)$ hitting a certain threshold, say $\vert \Delta w_{d}(t) \vert > T$ , where $T$ is specified by the system operator. 

Severity of ramp events can further be categorized by defining different ramp event thresholds. To describe the amplitude, or severity, of ramp events, an ordered set of thresholds from system operator can be specified as $T_{min} < T_{1} < T_{2} \ldots < T_{max}$. By comparing $\Delta w_{d}(t)$ to these thresholds, one can identify into which interval the ramp events will fall. 

A “class” can be assigned to each ramp event depending in which interval it falls. Given the chronological wind power time series data {$w_{d}(t)$} back from the time pair . Then the corresponding time series of wind power changes {$\Delta w_{d}(t)$} is readily obtained. 

The objective in this paper is to predict the classes of the ramp events at the future time interval from t+1 to t+S where S indicates the number of time intervals into the future; this is in fact a S-step ahead classification problem until the current time.
\section{Methodolgy}

\subsection{Gradient Boosted Regression Tree}
Gradient Boosted Regression Trees is an ensemble learning method that builds a set of decision trees in a greedy manner at training time and makes a class prediction by taking a majority vote, which is the mode of classes predicted by individual trees at test time. 

The model is an ensemble of k trees, formalized as
\begin{equation}
\hat{y}_{i} = \sum_{k=1}^{K} \ f_{k}(x_{i}), \ f_{k} \in  \ F
\end{equation}
The model is built in a greedy manner, with an aim to minimize the objective function 
\begin{equation}
\ Obj = \sum_{i}^{n} \ l(\hat{y}_{i},y) + \sum_{k} \Omega(f_{k}) 
\end{equation}

To find the functions $f_{k}$ , an additive training procedure is followed
\begin{align*}
\hat{y}_{i}^{(0)} &= 0 \\
\hat{y}_{i}^{(1)} &= \ f_{1}(x_{i}) = \hat{y}_{i}^{(0)} + f_{1}(x_{i}) \\
\hat{y}_{i}^{(2)} &= \ f_{1}(x_{i})+ f_{2}(x_{i}) = \hat{y}_{i}^{(1)} + f_{2}(x_{i}) \\
\ldots \\
\hat{y}_{i}^{(t)} &= \sum_{k=1}^{t} \ f_{k}(x_{i}) = \hat{y}_{i}^{(t-1)} + f_{t}(x_{i}) \\
\end{align*}

The model at training round $t$ is the sum of the function in the previous round and a function $f_{t}(x)$ . To find the function $f_{t}(x)$, the following objective function is minimized at round $t$,
\begin{equation}
\ Obj^{(t)} = \sum_{i=1}^{n} l(y_{i}, \hat{y}_{i}^{(t)}) + \sum_{i=1}^{n} \Omega (f_{i})
\end{equation}

A quadratic approximation allows us to define the objective function at round $t$ as
\begin{equation}
\ Obj^{(t)} \simeq \sum_{i=1}^{n} [g_{i} f_{t}(x_{i}) + \frac{1}{2}h_{i}f_{t}^{2}(x_{i}) ] +  \Omega (f_{t})
\end{equation}
where
\begin{equation}
g_{i} = {\partial}_{\hat{y}^(t-1)} l(y_{i},\hat{y}^{(t-1)}) , h_{i} = {\partial}_{\hat{y}^(t-1)}^{2} l(y_{i},\hat{y}^{(t-1)})
\end{equation}

Thus, the search for the functions $f_{t}$ depend on the objective only via the first and second gradients of the loss functions $g_{i}$ and $h_{i}$ respectively, which inspires the name Gradient Boosted Regression Trees.

This model is widely used in the industry for a variety of regression and classification problems, across a number of domains.

\section{Case studies}
\subsection{Dataset description}
The model validation is performed using wind power data obtained from the National Renewable Energy Laboratory (NREL) website~\cite{NREL}.
Modeled wind farm data points for the Eastern region of
United States are provided in the dataset. This database was developed for the purpose of assisting researchers involved in wind integration
studies. Wind speed/power data pertaining to 1326 onshore
sites in 34 states and 4948 offshore sites for 17 states is available in the dataset with a time resolution of 10 minutes. In the present study, we use data corresponding to one offshore site (SITE 7856) and one onshore site (SITE 13000). The wind speed and net power are listed as features over 157,969 time points in both the datasets.


In order to formulate the wind ramp event classification problem, it is necessary to tranform the numerical values of the wind power (targets of the training data sets) into labels or classes. Data is labelled according to certain thresholds which are fixed by the operators/ plant owners as per their requirements. To prepare the data, differences between wind power at successive 10-minute, 20- minute, 30-minute, 40-minute, 50-minute, 60-minute time points are taken. The obtained values are compared with the threshold set at 50 per cent of the rated site capacity, which is 10 MW for SITE 13000 and 545 MW for SITE 7856.\\
For example, using these threshold values, the targets are classified into four distinct classes in the following manner for the SITE 13000 dataset.

\begin{equation}
\begin{split}
  target~class=
    1,& ~ \text{if $x<-10$}.\\
    2,& ~ \text{if $-10<x<0$}.\\
    3,& ~ \text{if $0<x<10$}.\\
    4,& ~ \text{if $x>10$}.\\
    \end{split}
 \end{equation}

where x is the actual point value of the wind power difference between successive time points. The training data thus created is now used to train various classification models.

The class distribution for the datasets is highly imbalanced, as depicted in Tables I and II for SITE 7856 dataset and SITE 13000 dataset respectively. Class 1 and 4 represent severe down and up ramp events respectively, and thus are rare in number. It can be observed in Table I that only 0.58\% of the examples represent the rare ramp up/down events. This can also be observed in the SITE 13000 dataset depicted in Table II where only 5.06\% of the examples correspond to the rare events. The occurance of rare events is higher in site 13000 compared to site 7856 but still the fraction is too small for efficient traning. Classifying such events accurately is very crucial for enhancing the reliability and safe integration of wind power into the grid. 
\begin{table}[htbp]
\renewcommand{\arraystretch}{1.3}
\caption{SITE 7856 dataset}
\label{tab:example}
\centering
\begin{tabular}{c|c|c}
	\hline
    Class &  Number of examples & Percentage\\
    \hline
 	\hline
    1 & 447 & 0.28\\
    \hline
    2 & 81760 & 51.70\\
    \hline 
    3 & 75377 & 47.72\\
    \hline
    4 & 474 & 0.30\\
    \hline
\end{tabular}
\end{table}

\FloatBarrier

\begin{table}[h]
\renewcommand{\arraystretch}{1.3}
\caption{SITE 13000 dataset}
\label{tab:example}
\centering
\begin{tabular}{c|c|c}
	\hline
    Class &  Number of examples & Percentage\\
    \hline
 	\hline
    1 & 3611 & 2.29\\
    \hline
    2 & 83270 & 52.71\\
    \hline 
    3 & 66685 & 42.21\\
    \hline
    4 & 4402 & 2.77\\
    \hline
\end{tabular}
\end{table}
\FloatBarrier
Since predicting these rare ramp events is of greater value, we explicitly measure the performance of our models according to the F1 score for rare ramp event class (1 and 4). We see in section V that the GBRT model performs competitively according to this metric on SITE 13000 dataset.

\subsection{Model selection}

XGBoost is an open-source implementation of parallelized gradient boosted regression trees, which has been shown to work well in a variety of machine learning competitions across a wide variety of datasets. It supports various
objective functions, including regression, classification and ranking. The
package is made to be extensible, and is suited for research purposes. 
XGBoost requires careful parameter tuning to ensure optimal model performance.
The parameters \textit{nEstimators}, which controls the number of rounds for which boosting is performed, and \textit{maxDepth}, which is the maximum depth of a decision tree, were determined with a grid-search based approach. The basic idea behind grid search is to pick out the best combination of parameters from the grid formed by individual parameter values.
\textit{nEstimators} varied from [50,100,200], and \textit{maxDepth} varied from [2,4,6]. The best parameters were selected on the basis of the best 3-fold cross validation performance on the train set, to ensure better generalization to the test set.

\subsection{Results}
We benchmark the performance of a parallelized implementation of the GBRT model against a variety of state-of-the-art machine learning algorithms: Support Vector Machines(SVM), Neural Networks (NN), LSTM-NN(Long Short Term Memory)-Neural Networks and Random Forests(RF). 

The GBRT model performs robustly across different datasets, outperforming the other models in terms of both classification accuracy and mean F1 score. The classification accuracy is defined as follows.
\begin{equation}
Accuracy = \frac{n_{c}}{n_{t}}
\end{equation}
where $n_{c}$ is the number of correctly classified instances and $n_{t}$ is the total number of instances. 
 
The F1 score is the harmonic mean of precision and recall, defined as 
\begin{equation}
F_{1} = \frac{2*precision*recall}{precision+recall}
\end{equation}

where precision is defined as the fraction of true positives out of total instances predicted as positives, that is

\begin{equation}
Precision = \frac{t_{p}}{t_{p}+f_{p}}
\end{equation}

and recall is defined as the fraction of instances belonging to positive class that were predicted as positives, that is

\begin{equation}
Recall = \frac{t_{p}}{t_{p}+f_{n}}
\end{equation}

\subsection{Benchmark Models}
\begin{enumerate}

\item \textbf{Support Vector Machines (SVMs)} \newline
SVMs construct a separating hyperplane in a high-dimensional space, and have been applied to a wide variety of classification and regression problems. 
SVMs have been applied to the problem of wind ramp event forecasting~\cite{burges1998tutorial}~\cite{zeng2011support}. \newline
However, their applicability to large datasets like SITE 13000 and site 7856 is limited, as they take a long time to train and can't be parallelized trivially. Moreover, the kernel SVM implementation is rendered unfeasible as the kernel matrix takes up large memory space. The scores in Table 3 and Table 4 were obtained with a linear SVM implementation from scikit-learn Python library. 

\item \textbf{Artifical Neural Networks (ANNs)} \newline 
ANNs are a family of models inspired by biological neural networks and are used to estimate or approximate functions that can depend on a large number of inputs and are generally unknown.
ANN models have been used to forecast wind power and for wind ramp event class prediction~\cite{plaut1987learning}~\cite{liu2012wind}. Neural networks are a class of machine learning models that mimic the neurons in the human brain, and can be shown to learn arbitrarily complex mappings approximately. 

\item \textbf{Long Short Term Memory Networks (LSTMs)} \newline
LSTMs are a variant of Recurrent Neural Networks which contain LSTM blocks, with specialised input, forget and output gates and maintain a cell state, which maintains a "thought vector" summarizing the data seen so far.~\cite{greaves2009temporal}~\cite{hochreiter1997long}  \newline
Some studies have showcased the efficacy of Recurrent Neural Network(RNN) models with time-series data, for classification, regression and predictive modelling tasks. We benchmark the performance of the GBRT model against a variant of RNNs, called LSTM (Long-Short term memory). These models have been shown to work very well for sequence modelling, and are used for a variety of tasks including speech recognition, handwriting generation.  

\item \textbf{Random Forests} \newline
Random forests are an ensemble learning method for classification, regression and other tasks, that operate by constructing a multitude of decision trees at training time and outputting the class that is the mode of the classes (classification).~\cite{breiman2001random} \newline
Random forests are resistant to overfitting in general. Random Forests have been applied to the problem of wind power forecasting as well~\cite{linseasonal}.

\item \textbf{Persistence benchmark} \newline
Persistence prediction uses the ground truth ramp class in the previous time-step and predicts the same class for the current time-step. This serves as a baseline for advanced machine learning models to compare their performance.
\end{enumerate}

\subsubsection{Implementation details}
The GBRT model was implemented using open source machine learning library xgboost~\cite{xgboost}. The cross-validated grid search for the parameters was done using the xgboost's interface to Python's scikit-learn library. The ANN and LSTM models were trained on a Amazon Web Services(AWS) GPU node with 16GB RAM and a Tesla K20 GPU with 4 GB memory and 40 cores. The model was implememented using the open-source deep learning library Keras\cite{keras}. \\
The ANN model consists of an input layer of size 36, 2 hidden layers with 100 neurons each, with dropout probability of 0.2 in each layer and the final layer as a 4-way softmax layer. \\
The LSTM model consists of hidden layer with 36 timesteps, with the input being fed into each timestep being the wind power. The final layer is a 4-way softmax from the hidden layer of size 4 to the softmax output. \\
Both  models were trained using minibatch stochastic gradient descent with a minibatch size of size 512 for 20 epochs, with the loss function being the categorical cross-entropy loss with 4 categories. The data was split as 80 percent training and 20 percent test data using a stratified shuffle split strategy. The validation set was taken as 10 percent of the training data. The results are reported with the weights at the end of the $20^{th}$ epoch. \newline
The Random Forest model was implemented with Python's scikit-learn library \cite{scikit-learn}\cite{scikit} and cross-validated grid search was performed to select the parameters. 
\textit{nEstimators} varied from [50,100,200] and \textit{maxFeatures} varied between [sqrt(numFeatures),log2(numFeatures)], which is the maximum number of features to consider when looking for the best split.  3-fold cross validation was performed to select the best set of parameters.

\section{Results}

Table III and Table IV compare the classification accuracy, overall F1 score and F1 score for the rare ramp events for the GBRT model versus the benchmark models, for the SITE 13000 and SITE 7856 datasets. The overall F1 score is computed as the average of the F1 scores for class prediction for ramp event in the next 10, 20, 30, 40, 50 and 60 minute windows. The accuracy and F1 scores for the rare classes are computed in a similar fashion.\newline

Looking at the SITE 13000 dataset in Table III, the GBRT model achieves an overall F1 score of 0.83 and the Random Forest model achieves an overall F1 score of 0.79. Comparing this to the neural network based models with F1 scores of 0.74, the decision tree based models are better predictors of the class of a wind ramp event. The persistence benchmark performs very well with a F1 score of 0.77. \\
Looking at the accuracy of the models, GBRT gets an accuracy of 0.82 and the Random Forest model gets an accuracy of 0.80. Comparing this to LSTM-NNs and ANNs, which achieve an accuracy of 0.75, we see that the Random Forest and GBRT models obtain higher accuracy than the persistence benchmark while the LSTMs and ANNs perform below the benchmark set at 0.77 \\ 
For the SITE 7856 dataset in Table IV, the persistence benchmark forms a good baseline with a F1 score of 0.74 . The GBRT model achieves an overall F1 score of 0.79 . In comparison, the LSTM-NN and ANN models come up with an overall F1 score of 0.77 and 0.74 . The SVM and Random Forests model perform below the persistence benchmark at overall F1 scores of 0.63 and 0.53 respectively. \\
The accuracy of the models on SITE 7856 dataset is compared in Table IV. The GBRT gets an accuracy of 0.79, followed by ANNs and LSTMs, which get 0.78 and 0.75 respectively. The persistence benchmark is set at 0.74, which the Random Forest and SVM models perform below at 0.53 and 0.64 respectively. \\
In order for any such model to be practically applicable, it needs to have a reasonable F1 score for the severe up-ramp and down-ramp events. For the SITE 13000 dataset, the GBRT model has a rare event F1 score of 0.58. The ANN model achieves a score of 0.54 and the RF model achieves a score of 0.50. The persistence benchmark is set at 0.49, which the LSTM-NNs and SVMs perform below at 0.11 and 0.13 respectively. \\
For the SITE 7856 dataset, the ANN model gives a rare event F1 score of 0.50 followed by GBRT model with a score of 0.42. The persistence benchmark is set at 0.31, which the Random Forests, LSTM-NN and the SVM models fail to beat with a score of 0.15, 0.14 and 0.00 respectively. \\

In conclusion, performing a thorough analysis of the performance of various ML models on both datasets shows that the GBRT model robustly achieves high accuracy the other ML models across datasets. In particular, with a reasonably high F1 score for rare events in both datasets, the GBRT model is practically applicable to the problem of wind ramp event class prediction.

\begin{table}[h]
\renewcommand{\arraystretch}{1.3}
\caption{SITE 13000 dataset}
\label{tab:example}
\centering
\begin{tabular}{c|c|c}
	\hline
	\multicolumn{3}{c}{Gradient Boosted Regression Trees(GRBT) model}\\
    \hline
    Accuracy & F1 score(overall) & F1 score (rare events)\\
    \hline
    \hline
    \textbf{0.82} & \textbf{0.83} & \textbf{0.58}\\
    \hline
\end{tabular}
\begin{tabular}{c|c|c}
	\multicolumn{3}{c}{Support Vector Machine(SVM) model}\\
    \hline
    Accuracy & F1 score(overall) & F1 score (rare events)\\
    \hline
    \hline

    0.62 & 0.59 & 0.13\\
    \hline
\end{tabular}
\begin{tabular}{c|c|c}
	\multicolumn{3}{c}{Artificial Neural Network(ANN) model}\\
    \hline
    Accuracy & F1 score(overall) & F1 score (rare events)\\
    \hline
    \hline
    0.75 & 0.74 & 0.54 \\
    \hline
\end{tabular}
\begin{tabular}{c|c|c}
	\multicolumn{3}{c}{Long Short Term Memory-Neural Network (LSTM-NN) model}\\
    \hline
    Accuracy & F1 score(overall) & F1 score (rare events)\\
    \hline
    \hline
    0.75 & 0.74 & 0.11 \\
    \hline
\end{tabular}
\begin{tabular}{c|c|c}
	\multicolumn{3}{c}{Random Forests (RF) model}\\
    \hline
    Accuracy & F1 score(overall) & F1 score (rare events)\\
    \hline
    \hline
	0.79 & 0.80 & 0.50 \\
    \hline
\end{tabular}
\pagebreak
\begin{tabular}{c|c|c}
	\multicolumn{3}{c}{Persistence Benchmark}\\
    \hline
    Accuracy & F1 score(overall) & F1 score (rare events)\\
    \hline
    \hline
    0.77 & 0.77 & 0.49 \\
    \hline
\end{tabular}
\end{table}

\FloatBarrier


The testing times of the models are compared across different hardware setups as mentioned in the implementation details is compared in Table V. \\
The neural network based models are faster at test times because of the parallelism afforded by the GPU, as opposed to the decision-tree based models which only used a multi-core CPU for training. \\
Since the test times per example are much lower than the smallest time window for prediction(10 minutes), these models are suitable for real-time prediction in a streaming input setting.

\begin{table}[h]
\renewcommand{\arraystretch}{1.3}
\caption{SITE 7856 dataset}
\label{tab:example}
\centering
\begin{tabular}{c|c|c}
	\hline
	\multicolumn{3}{c}{Gradient Boosted Regression Trees(GBRT) model}\\
    \hline
    Accuracy & F1 score(overall) & F1 score (rare events)\\
    \hline
    \hline
     \textbf{0.79} & \textbf{0.79} & 0.42 \\
    \hline
    
\end{tabular}
\begin{tabular}{c|c|c}
    \multicolumn{3}{c}{Support Vector Machines(SVM) model}\\
    \hline
    Accuracy & F1 score(overall) & F1 score (rare events)\\
    \hline
    \hline
    0.64 & 0.63 & 0.00 \\
    \hline
\end{tabular}

\begin{tabular}{c|c|c}
	\multicolumn{3}{c}{Artifical Neural Network(ANN) model}\\
    \hline
    Accuracy & F1 score(overall) & F1 score (rare events)\\
    \hline
    \hline

    0.78 & 0.77 & \textbf{0.50} \\
    \hline
    
\end{tabular}
\begin{tabular}{c|c|c}
	\multicolumn{3}{c}{Long Short Term Memory(LSTM-NN) model}\\
    \hline
    Accuracy & F1 score(overall) & F1 score (rare events)\\
    \hline
    \hline

    0.75 & 0.74 & 0.14 \\
    \hline

\end{tabular}
\begin{tabular}{c|c|c}
	\multicolumn{3}{c}{Random Forest model}\\
    \hline
    Accuracy & F1 score(overall) & F1 score (rare events)\\
    \hline
    \hline

    0.53 & 0.53 & 0.15 \\
    \hline

\end{tabular}
\begin{tabular}{c|c|c}
	\multicolumn{3}{c}{Persistence}\\
    \hline
    Accuracy & F1 score(overall) & F1 score (rare events)\\
    \hline
    \hline

    0.74 & 0.74 & 0.31 \\
    \hline

\end{tabular}
\end{table}

\FloatBarrier

\begin{table}[h]
\renewcommand{\arraystretch}{1.3}
\caption{Test times per example in milliseconds}
\label{tab:example}
\centering
\begin{tabular}{c|c|c}
	\hline
    Model &  SITE 13000  & SITE 7856 \\
    \hline
 	\hline
    GBRT & 75 & 72.3\\
    \hline
    SVM & 81 & 82.5\\
    \hline 
    ANN & 60.3 & 70.6\\
    \hline
    LSTM-NN & 112.5 & 115.7\\
    \hline
    RF & 86.4 & 86.7\\
    \hline
    Persistence & 0.1 & 0.1\\
    \hline
\end{tabular}
\end{table}
\FloatBarrier

\section{Conclusion}
Gradient Boosted regression trees are a powerful class of models for learning patterns from a wide variety of datasets. They have been shown to be robust to overfitting, easy to train and compact to use at test time. This work shows that wind ramp event prediction from large historical datasets is made possible by parallelizing the learning procedure of gradient boosted trees across multi-core machines with a large number of cores. 
The performance of the GBRT model is compared to that of state-of-the-art pattern recognition and machine learning algorithms such as Support Vector Machines, Artificial Neural networks, Long Short Term Memory Neural Networks and Random Forests. The comparison is done on two datasets from wind farms in Alberta and the GBRT model is shown to work well on both datasets across three metrics. In addition to the regularly used metrics of accuracy and F1 score, we introduced an additional metric for F1 score of rare events to evaluate the practical usability of these models. The GBRT and ANN models perform well with respect to this metric.\\
We compare the performance of these models with a simple persistence benchmark. The superior performance of the persistence model to a number of machine learning models mentioned above indicates that the performance of wind ramp event prediction models may be improved by designing a models that take in the last persistent class as input along with the historical data to make predictions, a direction that can be explored in future work.
\ifCLASSOPTIONcaptionsoff
  \newpage
\fi

\section{ACKNOWLEDGEMENT}
This work was supported by the Scientists’ Pool Scheme
of the Council of Scientific and Industrial Research (CSIR),
Government of India (No. 8741-A). The third author would
like to acknowledge the financial support by the Centre for
Intelligent Systems Research (CISR) at Deakin University, Australia. 



%


\bibliography{Bib_bank.bib}{}

\begin{thebibliography}{10}
\providecommand{\url}[1]{#1}
\csname url@samestyle\endcsname
\providecommand{\newblock}{\relax}
\providecommand{\bibinfo}[2]{#2}
\providecommand{\BIBentrySTDinterwordspacing}{\spaceskip=0pt\relax}
\providecommand{\BIBentryALTinterwordstretchfactor}{4}
\providecommand{\BIBentryALTinterwordspacing}{\spaceskip=\fontdimen2\font plus
\BIBentryALTinterwordstretchfactor\fontdimen3\font minus
  \fontdimen4\font\relax}
\providecommand{\BIBforeignlanguage}[2]{{%
\expandafter\ifx\csname l@#1\endcsname\relax
\typeout{** WARNING: IEEEtran.bst: No hyphenation pattern has been}%
\typeout{** loaded for the language `#1'. Using the pattern for}%
\typeout{** the default language instead.}%
\else
\language=\csname l@#1\endcsname
\fi
#2}}
\providecommand{\BIBdecl}{\relax}
\BIBdecl

\bibitem{Bilgen2008372}
\BIBentryALTinterwordspacing
S.~Bilgen, S.~Keles, A.~Kaygusuz, A.~Sari, and K.~Kaygusuz, ``Global warming
  and renewable energy sources for sustainable development: A case study in
  turkey,'' \emph{Renewable and Sustainable Energy Reviews}, vol.~12, no.~2,
  pp. 372 -- 396, 2008. [Online]. Available:
  \url{http://www.sciencedirect.com/science/article/pii/S1364032106001183}
\BIBentrySTDinterwordspacing

\bibitem{Doe:2009:Online}
\BIBentryALTinterwordspacing
{\relax U.S. Department of Energy}. (2008, Jul.) 20 percent wind energy by
  2030: Increasing wind energy?s contribution to {U.S.} electricity supply
  doe/go-102008-2567, jul. 2008 {@ONLINE}. [Online]. Available:
  \url{http://www.nrel.gov/docs/fy08osti/41869.pdf}
\BIBentrySTDinterwordspacing

\bibitem{GWEO}
\BIBentryALTinterwordspacing
Global{\ }Wind{\ }Energy{\ }Council. (2014) Global wind energy outlook.
  [Online]. Available:
  \url{http://www.gwec.net/wp-content/uploads/2014/10/GWEO\_WEB.pdf}
\BIBentrySTDinterwordspacing

\bibitem{MNRE}
\BIBentryALTinterwordspacing
M.~of~New and R.~E.~G. of~India. (2015, May) Physical progress (achievements).
  [Online]. Available:
  \url{http://www.mnre.gov.in/mission-and-vision-2/achievements/}
\BIBentrySTDinterwordspacing

\bibitem{WE:WE96}
\BIBentryALTinterwordspacing
L.~Landberg, G.~Giebel, H.~A. Nielsen, T.~Nielsen, and H.~Madsen, ``Short-term
  prediction: An overview,'' \emph{Wind Energy}, vol.~6, no.~3, pp. 273--280,
  2003. [Online]. Available: \url{http://dx.doi.org/10.1002/we.96}
\BIBentrySTDinterwordspacing

\bibitem{Lange}
M.~Lange and U.~Focken, \emph{Physical Approach to Short-term Wind Power
  Prediction}.\hskip 1em plus 0.5em minus 0.4em\relax Heidelberg, Germany:
  Springer-Verlag, 2005.

\bibitem{Suganthan}
Y.~Ren, P.~Suganthan, and N.~Srikanth, ``A comparative study of empirical mode
  decomposition-based short-term wind speed forecasting methods,'' \emph{IEEE
  Transactions on Sustainable Energy}, vol.~6, no.~1, pp. 236--244, Jan 2015.

\bibitem{Mohandes2004939}
M.~Mohandes, T.~Halawani, S.~Rehman, and A.~A. Hussain, ``Support vector
  machines for wind speed prediction,'' \emph{Renewable Energy}, vol.~29,
  no.~6, pp. 939 -- 947, 2004.

\bibitem{Hui2012545}
\BIBentryALTinterwordspacing
H.~Liu, C.~Chen, H.~qi~Tian, and Y.~fei Li, ``A hybrid model for wind speed
  prediction using empirical mode decomposition and artificial neural
  networks,'' \emph{Renewable Energy}, vol.~48, no.~0, pp. 545 -- 556, 2012.
  [Online]. Available:
  \url{http://www.sciencedirect.com/science/article/pii/S096014811200362X}
\BIBentrySTDinterwordspacing

\bibitem{Sfetsos200023}
A.~Sfetsos, ``A comparison of various forecasting techniques applied to mean
  hourly wind speed time series,'' \emph{Renewable Energy}, vol.~21, no.~1, pp.
  23--35, 2000.

\bibitem{Pinson2007}
\BIBentryALTinterwordspacing
P.~Pinson, H.~A. Nielsen, J.~K. M?ller, H.~Madsen, and G.~N. Kariniotakis,
  ``Non-parametric probabilistic forecasts of wind power: required properties
  and evaluation,'' \emph{Wind Energy}, vol.~10, no.~6, pp. 497--516, 2007.
  [Online]. Available: \url{http://dx.doi.org/10.1002/we.230}
\BIBentrySTDinterwordspacing

\bibitem{Khosravi2013}
A.~Khosravi, S.~Nahavandi, and D.~Creighton, ``Prediction intervals for
  short-term wind farm power generation forecasts,'' \emph{IEEE Transactions on
  Sustainable Energy}, vol.~4, no.~3, pp. 602--610, July 2013.

\bibitem{Bludszuweit2008}
H.~Bludszuweit, J.~Dominguez-Navarro, and A.~Llombart, ``Statistical analysis
  of wind power forecast error,'' \emph{IEEE Transactions on Power Systems},
  vol.~23, no.~3, pp. 983--991, Aug 2008.

\bibitem{Argonne}
C.~Monteiro, R.~Bessa, V.~Miranda, A.~Botterud, J.~Wang, and G.~Conzelmann,
  ``Wind power forecasting: State-of-the-art 2009,'' Argonne Nat. Lab., Tech.
  Rep. ANL/DIS-10-1., November 2009.

\bibitem{Bremnes2006}
\BIBentryALTinterwordspacing
J.~B. Bremnes, ``A comparison of a few statistical models for making quantile
  wind power forecasts,'' \emph{Wind Energy}, vol.~9, no. 1-2, pp. 3--11, 2006.
  [Online]. Available: \url{http://dx.doi.org/10.1002/we.182}
\BIBentrySTDinterwordspacing

\bibitem{sevlian6555972}
R.~Sevlian and R.~Rajagopal, ``Detection and statistics of wind power ramps,''
  \emph{IEEE Transactions on Power Systems}, vol.~28, no.~4, pp. 3610--3620,
  Nov 2013.

\bibitem{bossavy2010forecasting}
A.~Bossavy, R.~Girard, and G.~Kariniotakis, ``Forecasting uncertainty related
  to ramps of wind power production,'' in \emph{European Wind Energy Conference
  and Exhibition 2010, EWEC 2010}, vol.~2.\hskip 1em plus 0.5em minus
  0.4em\relax European Wind Energy Association, 2010, pp. 9--pages.

\bibitem{greaves2009temporal}
B.~Greaves, J.~Collins, J.~Parkes, and A.~Tindal, ``Temporal forecast
  uncertainty for ramp events,'' \emph{Wind Engineering}, vol.~33, no.~4, pp.
  309--319, 2009.

\bibitem{Mingjian7018976}
M.~Cui, D.~Ke, Y.~Sun, D.~Gan, J.~Zhang, and B.-M. Hodge, ``Wind power ramp
  event forecasting using a stochastic scenario generation method,'' \emph{IEEE
  Transactions on Sustainable Energy}, vol.~6, no.~2, pp. 422--433, April 2015.

\bibitem{Zarei6039625}
H.~Zareipour, D.~Huang, and W.~Rosehart, ``Wind power ramp events
  classification and forecasting: A data mining approach,'' in \emph{IEEE Power
  and Energy Society General Meeting, 2011}, July 2011, pp. 1--3.

\bibitem{NREL}
\BIBentryALTinterwordspacing
N.~R.~E. Laboratory. (2015, Feb.) http://www.nrel.gov/wind/ {@ONLINE}.
  [Online]. Available:
  \url{http://www.nrel.gov/wind/integrationdatasets/eastern/data.html}
\BIBentrySTDinterwordspacing

\bibitem{burges1998tutorial}
C.~J. Burges, ``A tutorial on support vector machines for pattern
  recognition,'' \emph{Data mining and knowledge discovery}, vol.~2, no.~2, pp.
  121--167, 1998.

\bibitem{zeng2011support}
J.~Zeng and W.~Qiao, ``Support vector machine-based short-term wind power
  forecasting,'' in \emph{Power Systems Conference and Exposition (PSCE), 2011
  IEEE/PES}.\hskip 1em plus 0.5em minus 0.4em\relax IEEE, 2011, pp. 1--8.

\bibitem{plaut1987learning}
D.~C. Plaut and G.~E. Hinton, ``Learning sets of filters using
  back-propagation,'' \emph{Computer Speech \& Language}, vol.~2, no.~1, pp.
  35--61, 1987.

\bibitem{liu2012wind}
Z.~Liu, W.~Gao, Y.-H. Wan, and E.~Muljadi, ``Wind power plant prediction by
  using neural networks,'' in \emph{Energy Conversion Congress and Exposition
  (ECCE), 2012 IEEE}.\hskip 1em plus 0.5em minus 0.4em\relax IEEE, 2012, pp.
  3154--3160.

\bibitem{hochreiter1997long}
S.~Hochreiter and J.~Schmidhuber, ``Long short-term memory,'' \emph{Neural
  computation}, vol.~9, no.~8, pp. 1735--1780, 1997.

\bibitem{breiman2001random}
L.~Breiman, ``Random forests,'' \emph{Machine learning}, vol.~45, no.~1, pp.
  5--32, 2001.

\bibitem{linseasonal}
Y.~Lin, U.~Kruger, J.~Zhang, Q.~Wang, L.~Lamont, and L.~E. Chaar, ``Seasonal
  analysis and prediction of wind energy using random forests and arx model
  structures.''

\bibitem{xgboost}
\BIBentryALTinterwordspacing
Large-scale and distributed gradient boosting library. [Online]. Available:
  \url{https://github.com/dmlc/xgboost}
\BIBentrySTDinterwordspacing

\bibitem{keras}
\BIBentryALTinterwordspacing
Deep learning library for python. [Online]. Available:
  \url{https://github.com/fchollet/keras}
\BIBentrySTDinterwordspacing

\bibitem{scikit-learn}
F.~Pedregosa, G.~Varoquaux, A.~Gramfort, V.~Michel, B.~Thirion, O.~Grisel,
  M.~Blondel, P.~Prettenhofer, R.~Weiss, V.~Dubourg, J.~Vanderplas, A.~Passos,
  D.~Cournapeau, M.~Brucher, M.~Perrot, and E.~Duchesnay, ``Scikit-learn:
  Machine learning in {P}ython,'' \emph{Journal of Machine Learning Research},
  vol.~12, pp. 2825--2830, 2011.

\bibitem{scikit}
\BIBentryALTinterwordspacing
Machine learning library in python. [Online]. Available:
  \url{http://scikit-learn.org/stable/}
\BIBentrySTDinterwordspacing

\end{thebibliography}
\bibliographystyle{IEEEtran}




%









\end{document}